\documentclass[runningheads]{llncs}
\usepackage{graphicx}
\usepackage{listings}
\begin{document}
\title{From Probabilistic Programming to Complexity-based Programming}
%
%
\author{Giovanni Sileno\inst{1} \and
Jean-Louis Dessalles\inst{2}}
\authorrunning{G. Sileno and J.-L. Dessalles}
%
\institute{University of Amsterdam, Amsterdam, The Netherlands 
\and
Télécom Paris, Paris-Saclay University, France\\
\email{g.sileno@uva.nl}, \email{dessalles@telecom-paris.fr}}
\maketitle              
\begin{abstract}
The paper presents the main characteristics and a preliminary implementation of a novel computational framework named \textsc{CompLog}. Inspired by probabilistic programming systems like ProbLog, CompLog builds upon the inferential mechanisms proposed by Simplicity Theory, relying on the computation of two Kolmogorov complexities (here implemented as min-path searches via ASP programs) rather than probabilistic inference. The proposed system enables users to compute \textit{ex-post} and \textit{ex-ante} measures of unexpectedness of a certain situation, mapping respectively to posterior and prior subjective probabilities. The computation is based on the specification of world and mental models by means of causal and descriptive relations between predicates weighted by complexity. The paper illustrates a few examples of application: generating relevant descriptions, and providing alternative approaches to disjunction and to negation. 
\end{abstract}
\section{Introduction}
Probabilistic forms of programming are increasingly attracting attention \cite{DeRaedt2007,Saad2019,Hahn2022,Winters2022}, and part of this success is likely because they provide an intuitive common ground between numeric and symbolic approaches. Yet, the axiomatization of probability theory relies on assuming the presence of a measurable space of events, which can in principle be contested. Given any description of the world, we can always add new elements, eg. entities previously not in focus, arrangements at other moments in time,  properties not considered before. On the other hand, scalability concerns 
generally hinder considering an ever-growing number of variables. 

In previous work \cite{Sileno2021}, we have argued that the notion of ``unexpectedness'' introduced by Simplicity Theory (ST) \cite{Dessalles2008}, based on the computation of two bounded Kolmogorov complexities, provides a framework more general than Bayes' rule to compute the posterior of a certain situation. On the basis of this conjecture, we  propose here a novel computational framework named \textsc{CompLog}, able to compute \textit{ex-post} and \textit{ex-ante} measures of unexpectedness of a certain situation, mapping respectively to posterior and prior subjective probabilities. The computation relies on the representation of causal and descriptive relationships, specified as relationships between predicates, weighted by complexity. 

At syntactic level, the system we propose is inspired by {ProbLog} \cite{DeRaedt2007}, yet it has its own unique characteristics. (1) {CompLog} enables by design the distinction between descriptive and causal dimensions, and thus in principle supports the integration of distinct dedicated tools for the two dimensions (ontologies, and causal models, for instance). (2) With adequate operators, CompLog can mimic cognitive mechanisms observable in humans. This is because the Simplicity Theory of cognition on which it builds upon has a better alignment with evaluations of relevance expressed by humans compared to Shannon's Information theory. (3) From a technical point of view, because the minimization of complexity is based upon a non-extensional search (the framework does not keep and propagate random distributions), we hypothesize that {CompLog} may be faster than a probabilistic equivalent system, and thus in principle it may be used efficiently both for abduction and prediction (causal, non-primarily statistical). 

The present paper does not aim to demonstrate all these goals, but will present and discuss the first design choices we are taking in this line of work. Section 2 provides a brief overview to Simplicity Theory, the cognitive model underlying the inferential mechanisms we implement; on {ProbLog}; and on some of the problems lying at the boundary between logical and probabilistic approaches. Section 3 presents the main {CompLog}'s components: a (simplistic) programming language, inferential mechanisms, and a preliminary implementation in ASP. Section 4 provides a few examples of application, meant to show CompLog's distinctive characteristics in comparison with functionally comparable systems. 

\section{Theoretical Background}

\subsection{Simplicity Theory}

Shannon's theory of information 
entails that samples extracted from a uniform distribution are maximally informative. Yet, very few humans would agree with such a conclusion. Motivated by addressing this shortcoming, Simplicity Theory (ST) \cite{Dessalles2008} was presented as an alternative computational model of cognition. Technically, ST introduces a measure of \textit{unexpectedness} \cite{Dessalles2013} building upon results from \textit{algorithmic information theory}, which is defined as:
$$U(s) = C_W(s) - C_D(s)$$
where $s$ is a situation in focus, and $C_W$ and $C_D$ are two Kolmogorov complexities computed via distinct bounded Turing machines.\footnote{The \textit{Kolmogorov complexity} of a string $x$ is defined as the minimal length of a program that, given a certain optional input parameter $y$, produces $x$ as an output: $K_\phi(x|y) = \min_p \big\{|p| : p(y) = x \big\}$.
Note that the length of the minimal program depends on the operators and symbols available to the machine $\phi$.} 
The \textit{causal complexity} $C_W$ (also called \textit{world complexity} or \textit{generation complexity}) relies on a ``world model'' maintained by a world machine $W$, whose operators are expected to be about occurrences, causal dependencies, and possibly forms of causal compositionality. The \textit{description complexity} $C_D$ builds upon a ``mind model'' maintained by a description machine $D$, whose operators are expected to be about concept retrieval, association, and possibly various forms of descriptive compositionality. Informally, ST entails that situations are unexpected if they are (algorithmically) simpler to describe (ie. visualize mentally) than to generate in the world (ie. to simulate their generation according to the agent's world model). Because complexities are expressed on a logarithmic  scale, one may introduce an additional constraint:
$$ U(s) \ge 0 $$ 
to capture a principle of \textit{cognitive economy}: situations are described up to the extent they are unexpected to occur.

Previous works \cite{Dessalles2008,Dessalles2011,Dessalles2011a,Dessalles2013} have shown that this definition predicts various human phenomena observed in experimental settings, amongst which judgments on remarkable lottery draws (e.g. $11111$ is more unexpected than $64178$, even if the lottery is fair), coincidence effects (e.g. meeting by chance a friend in a foreign city is more unexpected than meeting any unknown person equally improbable), deterministic yet unexpected events (e.g. lunar eclipses), and many others. 

\subsubsection{Unexpectedness and Bayes' rule}

Interestingly, we have recently acknowledged  \cite{Sileno2021} that Bayes' rule can be seen as a specific instantiation of ST's Unexpectedness that: (a) makes a candidate ``cause'' explicit and does not select automatically the best candidate; (b) takes a frequentist-like approach for encoding observables. We traced therefore the following correspondence: 
$$U(s) = \min_c \overbrace{ \left[ C_W(c * s)  - C_D(s) \right]}^{\textrm{posterior}} = \min_c [ \overbrace{C_W(s||c)}^{\textrm{likelihood}} + \overbrace{C_W(c)}^{\textrm{prior}}  - \hspace{-3pt} \overbrace{C_D(s)}^{\textrm{evidence}} \hspace{-3pt}]$$
where ``$*$'' is a (temporal) sequential compositional operator, $c$ is a situation (candidate cause of $s$), and ``$||$'' map to the ``$|$'' notation in conditional probability, with an additional explicit temporal constraint (in this case $c$ has to occur before than $s$). 
Taking unexpectedness as an alternative measure of subjective posterior probability, it could in principle replace Bayesian inference in a probabilistic deduction system.



\subsection{ProbLog}

ProbLog \cite{DeRaedt2007} is plausibly the best known amongst the available solutions for probabilistic programming. It provides a suite of efficient algorithms for various inference tasks, relying on a conversion of a program, queries, and evidence to a weighted Boolean formula, and thus transforming the inferential task into weighted model counting, which can be solved using state-of-the-art methods. The knowledge bases of ProbLog programs can be represented in various ways, amongst which Prolog/Datalog facts.
Here is an example of code:

\begin{lstlisting}
0.5 :: friendof(john, mary).
0.5 :: friendof(mary, pedro).
0.5 :: friendof(mary, tom).
0.5 :: friendof(pedro, tom).

1.0 :: likes(X, Y) :- friendof(X, Y).
0.8 :: likes(X, Y) :- friendof(X, Z), likes(Z, Y).

evidence(likes(mary, tom)).
query(likes(mary, pedro)).
\end{lstlisting}

A ProbLog program consists of facts (eg. \texttt{friendof(john, mary)}), rules (eg. \texttt{likes(X, Y) :- friendof(X, Y)}), possibly annotated with a certain probability (eg. \texttt{0.5}). Special predicates are used to specify goals (\texttt{query/1}) and  observations (\texttt{evidence/1}). 
ProbLog also introduces the possibility to specify disjoint facts in the head of rules to capture the existence of mutually exclusive outcomes.
Solving the previous program\footnote{See eg. \url{\texttt{https://dtai.cs.kuleuven.be/problog/editor.html}} for an online running version of ProbLog.} returns as outcome \texttt{probability: 0.58333333}.

\subsection{From ProbLog to CompLog}

The outcome of the ProbLog program above can be read both as saying that Mary likes Pedro a bit, or that Mary likes Pedro in more than half of the possible world configurations specified by the program.
Indeed, applications of probability theory oftentimes are used to deal with two dimensions of uncertainty: descriptive uncertainty (also \textit{indetermination}) and causal uncertainty. This is based on the assumption that the \textit{degree} or the extent to which it is true that, for instance, the email just received is spam may also be captured as the probability that this email is spam.  Whether and when the passage between the two types of uncertainties is legitimate is still an open question, which goes beyond probability theory and involves research fields such as fuzzy logic and other quantitative approaches to logic. Here we will consider however an alternative line of arguments with respect to this issue.

\subsubsection{Epistemic vs ontological}

On a more fundamental level, the descriptive dimension can be associated to \textit{epistemic uncertainty}, ascribed to things that have not been unveiled yet (or proven, in a logical framing), including things that may not be proven; it is therefore primarily a matter of \textit{conditions holding} in the world. In contrast, the causal dimension can be associated to \textit{ontological uncertainty}, ascribed to things that have not been created yet (or extracted, in a sampling framing); it is a matter of \textit{events occurring} in the world. Interestingly, seen in terms of conditions/events, passing from one dimension to the other can be related to various, and very distinct, challenges studied in AI and related fields. We will cover two prototypical cases here: symbolic automated reasoning about events, and causality in Bayesian networks.

\paragraph{Reasoning about events with logic programs}
In symbolic AI, particularly in the 80s/90s, reasoning about the effects of events (including actions) has been a mainstream topic, eventually related to the infamous \textit{frame problem}. It became soon manifest that it was rather difficult to perform automated reasoning about events by means of simple deduction. In order to deal with phenomena as inertia and locality of effects, several axiomatizations were proposed, the most known being \textit{situation calculus} \cite{McCarthy1969,Reiter2001}, \textit{event calculus} \cite{Kowalski1986,Shanahan1999}, \textit{fluent calculus}  \cite{Thielscher1999}. Looking at the event calculus, for instance, meta-level predicates are introduced to deal with two different types of entities: \textit{fluents} (ie. conditions that vary in time) and \textit{events} (ie. transitions/changes of fluents that occur at a certain point in time). Facts about the world are then reified either via \texttt{holds(C, T)} for fluents, or as \texttt{occurs(E, T)} for events, where \texttt{T} is a temporal coordinate.  Other solutions share similar meta-level constructs.

\paragraph{Interventions with Bayesian Networks}
In probabilistic inference, from a formal point of view, Bayesian Networks are known to capture only associationistic relationships. Any causal reading exists only in the mind of the modeler. To take into account interventions, like those present in causal scenarios (and relevant eg. to scientific experimentation), we need to consider an additional \texttt{do} operator \cite{Pearl2009,Pearl2016}. The \texttt{do} operator provides \textit{local counterfactuality} by performing an operation on the Bayesian network: its presence entails that the edges from the intervened node towards its parent nodes have to be cut. Because this operation is performed at the level of the Bayesian network, and not at the level of the variables, the construct can be seen also in this case as operating at a meta-level.

\subsubsection{Simplicity Theory as an integrative framework}
The examples above illustrate that: (1) both logical and probabilistic inferences work primarily at the level of conditions; and (2) additional machinery is required to properly take into consideration the side effects of events. ST, in contrast, explicitly relies on two different machines to compute unexpectedness: one for the world (making events occurring), and one for the mind (determining conditions). Intuitively, this cognitive model offers a more principled solution to separate ontological and epistemic uncertainty concerns, and indeed {CompLog} (in contrast eg. to PropLog or other probabilistic inference systems) stems from the idea of keeping them distinct.

\section{CompLog}
This section presents our current design choices on {CompLog} going over three dimensions: the programming language, the inferential mechanisms, and their implementation.

\subsection{Language}
We introduce here a simplistic propositional language meant to specify both causal and associative (descriptive) relationships. We do not require it to be comprehensive, but rather to be functional for early experimentations.

\paragraph{Conditions and events}
The language is based upon a primary distinction between conditions and events. Events are predicates whose duration of applicability, in the descriptive frame considered, is irrelevant, whereas conditions are predicates that can be applied for some relevant amount of time. We will then use simple literals to specify conditions: \texttt{s}, \texttt{x}, \texttt{y}, \texttt{z}. We will distinguish events from conditions by means of prefixed literals: \texttt{\#x}, \texttt{\#y}, etc. Two special type of events will be introduced, specified by their effects: \texttt{+x} for the initiation (or creation, production, addition, etc.) of condition \texttt{x}, \texttt{-x} for the termination (or destruction, consumption, removal, etc.) of a certain condition \texttt{x}.  

\vspace{-5pt}
\paragraph{Declarative and active rules}
A \textit{declarative rule} (to be read as \textit{if condition then conclusion}) is specified as a relation between two conditions, eg. \texttt{x -> y}. In the prolog/ASP syntax this construct would correspond to \texttt{y :- x}. An \textit{active rule} (to be read as \textit{if antecedent then consequent}) is specified as a relation between two events, eg. \texttt{\#push => +light} (if you push the button, the light goes on). 
Note that the syntax is sufficiently rich to describe causal dependencies with consumption (eg. \texttt{+x => +y, -x}) and without (eg. \texttt{+x => +y}).
Active rules can be also extended to fit the ECA template (\textit{when event in condition then action}): \texttt{\#push : electricity => +light}. 

\vspace{-5pt}
\paragraph{Race conditions}
The term ``race condition'' is used in concurrent systems to indicate a critical point in which only a limited amount of threads or processes have access to a resource, therefore execution may bring non-deterministic results, unless priority of access is defined up-front. Active rules in principle are susceptible to race conditions. For instance, given the following program:
\begin{lstlisting}
+x => +y.
+x => +z.
\end{lstlisting}
it is not clear whether only one amongst of \texttt{+y} or \texttt{+z} should be triggered after \texttt{+x}, or both of them. To make the syntax non-ambiguous, we assume that the state of the world changes always at the activation of any causal rule\footnote{There will be certainly a change of temporal coordinate, even if implicit.}, therefore the previous rules are interpreted as in a race condition, and only one of the two will be executed. To specify the absence of race conditions we can use the ECA template, allowing active rules with no triggering event, and introducing a \textit{catalyst} entity which is not consumed: 
\begin{lstlisting}
=> +x.
: x => +y.
: x => +z.
\end{lstlisting}
Active formulas with no triggering event can be seen as related to ergodic properties of the world, underlying asymptotic growth phenomena. In a propositional setting, growth is however constrained to be non-linear and capped. 

\vspace{-5pt}
\paragraph{Program}
A {CompLog} program consists of a \textit{world model}, specified as a set of causal relationships (in the proposed language, active rules) and centered on events; and a \textit{mental model}, specified as a set of associationistic relationships (eg. logical, in the proposed language via declarative rules), centered on conditions. If a program consists only of facts about events and active rules, it is called \textit{active program}. If a program consists only of facts about conditions and declarative rules it is called \textit{declarative program}. Any program can be divided into active and declarative programs, mapping isomorphically to world and to mental models.

\vspace{-5pt}
\paragraph{Program augmentation}
Given a declarative program, we can read it as an active program by capturing explicitly initiation and termination mechanisms. Here we will focus  only on initiation, as it has a more prominent role when thinking of the plausibility of a target situation. The transformation can be applied in two ways, with race conditions, or with no race conditions:
\begin{lstlisting}
x.  x -> y.    % declarative 
+x. +x => +y.  % active with race conditions
+x. : x => +y. % active without race conditions
\end{lstlisting}
\vspace{-5pt}
\paragraph{Complexities}
Whereas in {ProbLog} facts and rules can be given a probability (a value between 0 and 1), in {CompLog} facts and rules can be given a value of complexity (a positive value). When complexity is specified within a world model, associated to an element of an active program, it has to be interpreted as a causal complexity $C_W$, when it is specified within a mental model, associated to an element of a declarative program, it has to be interpreted as a descriptive complexity $C_D$. Consider the following example:
\begin{lstlisting}
4 :: eagle. 
12 :: #eagle. 
\end{lstlisting}
This program expresses that conceptually evoking an eagle is much easier than actually encountering it.\footnote{The easiness of retrieval may be due to a much more frequent appearance of the concept of eagle in discourses around the observer.} 

\subsection{Inference}

A second intuition motivating the introduction of {CompLog} is that the world and mental models connect to two different characterizations of computation, which can be treated separately. 

\vspace{-5pt}
\paragraph{Productive vs epistemic characterization}
Causal events generally consume resources (except for catalysts); we will therefore consider a \textit{productive} characterization of computation to simulate the world generation processes, which can be put in correspondence with graphical notations as, eg., Petri Nets. We will instead consider an \textit{epistemic} characterization for the computation of mental objects: once a fact has been unveiled (eg. it has been proven true or retrieved from memory), it becomes part of the current resources and cannot be removed, unless non-monotonic effects apply. This view can be applied to systems based on logic, and can also be given a graphical representation, as for instance through dependency graphs.

\paragraph{Computation as a ``colouring'' task}

From the above, we hypothesize that the world and mental models can be represented as two distinct directed graphs. The nodes coloured at a certain moment in time count as resources which are available at that moment. The edges reify the complexity, ie. the cost of colouring the target node, when the source node is coloured already. Queries specify goal nodes. With this mapping, the minimization of complexity required by the computation of Kolmogorov complexity maps to the application of min-path search algorithms. We require two distinct search algorithms, because transitions on the two graphs behave differently (respectively, with or without consumption). Consider for instance the following declarative program (mental model): 
\begin{lstlisting}
4 :: x. 4 :: y. 4 :: z.
1 :: z -> x. 1 :: z -> y.
3 :: x -> y. 3 :: y -> x.
\end{lstlisting}
Adding an initial state, the various relations can be represented on a graph like the one in Fig.~1.
We can then augment the declarative program to generate a corresponding active program, obtaining:
\begin{lstlisting}
4 :: +x. 4 :: +y. 4 :: +z.
1 :: +z => +x. 1 :: +z => +y. 
3 :: +x => +y. 3 :: +y => +x.
\end{lstlisting}
\begin{figure}[t]
    \centering
    \resizebox{6cm}{!}{\includegraphics{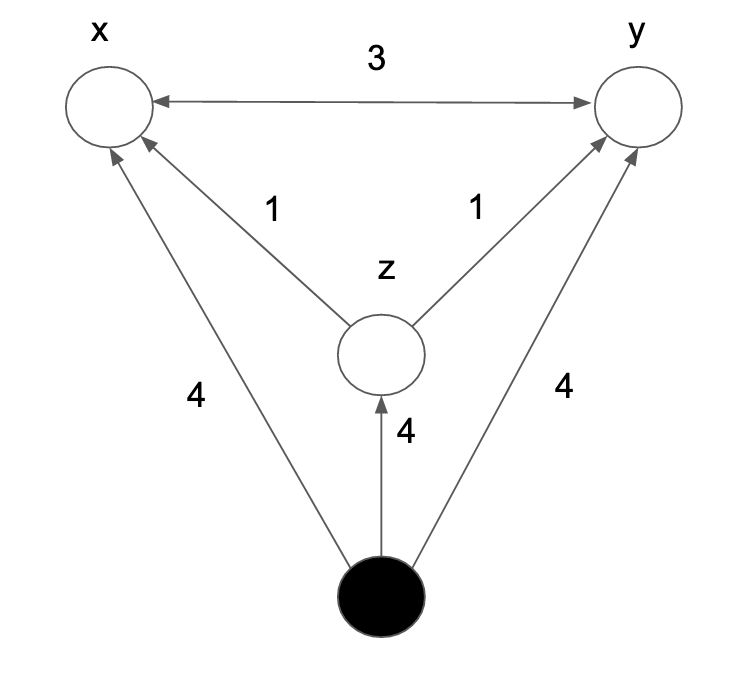}} \vspace{-10pt}
    \caption{Graph constructed from of a descriptive model. Coloured nodes (in black) are available resources. Numbers on edges specify the complexity of colouring the target node if the source node is coloured.}
    \label{fig:enter-label}
\vspace{-5pt}
\end{figure}
Note how we have illustratively kept the same primitive values of complexity in the two models, and yet the computed derived complexities vary. For instance, $C_D(\texttt{x}) = 4$ is equal to $C_W(\texttt{+x}) = 4$, whereas $C_D(\langle \texttt{x}, \texttt{y} \rangle) = 6$ is inferior to $C_W(\langle \texttt{+x}, \texttt{+y} \rangle) = 7$.\footnote{$\langle \ldots, \ldots \rangle$ is an operation used to specify multiple goals, all of which need to achieved, regardless of the order.} This divergence appears because on the descriptive program we apply an epistemic computation, which does not consume any resources; we can then generate \texttt{z} as a more efficient basis to construct both \texttt{x} and \texttt{y}. In contrast, on the active program we apply a productive computation, which (generally) consumes resources; the given program has a race condition over \texttt{+z}, and therefore it cannot be used both to trigger \texttt{+x} and \texttt{+y}. Note however how the designer can modify this behaviour in case \texttt{z} acts rather as a catalyst.

\paragraph{Output} 

In {CompLog}, given a query, we will compute its unexpectedness (mapping to a posterior, \textit{ex-post} probability) as the difference between the complexities computed on the world and on the description machine, ie. $U = C_W - C_D$. In the example illustrated in Fig.~1, we can infer for instance that $U(\langle \texttt{x}, \texttt{y} \rangle) = 1$, and therefore, this input would be (slightly) unexpected \textit{if it had to occur}. 

In order to generate an \textit{ex-ante} value, relevant for predicting what will occur in the future, the determination cost needs to be added back to unexpectedness, ie. $C_W^U = U + C_D$ \cite{Sileno2021}. On simple cases, this seems rather inefficient, as we could simply compute $C_W$. However, in general cases, there may be elements in the query (which describes a situation) which may turn out to be irrelevant. The optimization on $C_D$ to compute $U$ is functional to finding the correct framing by which to interpret the input situation. In probabilistic approaches, this is left implicit in the choice of random variables; for instance, we approach a lottery without considering the context that brought the lottery to be, nor other events that may disrupt its functioning: the framing is given by design.  For further information, see the \textit{informational principle of framing} elaborated in \cite{Sileno2021}.

\subsection{Implementation}

Because it relies on the computation of (bounded) Kolmogorov complexities, the core  inferential mechanism of CompLog are min-path search algorithms to be applied on the graphs resulting from the world and mental models. For our preliminary experiments, we have manually implemented the search algorithms in \textit{answer set programming} 
(ASP) \cite{Lifschitz2008}, using the ASP solver \texttt{clingo} \cite{Gebser2019}.\footnote{\url{\texttt{https://potassco.org/clingo/}}} This choice was motivated by the intention of keeping control on the various steps. Yet, other solutions may be more efficient.\footnote{Interestingly, a new probabilistic programming framework (including ProbLog inferential mechanisms) based on ASP/clingo has been recently presented: \texttt{plingo} \cite{Hahn2022}.}

The ASP program we use to encode the inferential mechanisms of {CompLog} consists of model and control parts, related respectively to the given problem (specific)  and to the control mechanisms (generic) required to run the inference.

The model part consists of statements reifying the edges of the graph with their cost (complexity), including the initial state (the coloured nodes) and the query (the goal nodes). The graph in Fig.~1 is for instance reified as:
\begin{lstlisting}
cost(s, x, 4). cost(s, y, 4). cost(s, z, 4).
cost(z, x, 1). cost(z, y, 1).
cost(x, y, 3). cost(y, x, 3).
start(s). goal(x). goal(y).
\end{lstlisting}

The control part of the program is an axiomatization consisting of three components: a common \textit{exploration} core, specific \textit{constraints} depending on whether we are in the context of productive or epistemic computation (ie. consumption or no-consumption, race conditions), and axioms for defining {optimization}. For the exploration axioms, we consider:
\begin{lstlisting}
path(X, Y) :- reached(X, N), edge(X, Y).
reached(X, 0) :- start(X).
:- goal(Y), not reached(Y, _).
\end{lstlisting}
The first line states that every outgoing edge from a reached node may be traveled. The second specifies that the initial state consists of nodes reached at time 0. The third states that all goals should be reached eventually. The search mechanism is then extended with specific constraints. In an epistemic computation setting, time is deemed irrelevant:
\begin{lstlisting}
{ reached(Y, N) } :- path(X, Y), reached(X, N).
\end{lstlisting}
which states that if the node \texttt{X} was reached at time \texttt{N}, and there is a path between \texttt{X} and \texttt{Y}, then \texttt{Y} may be reached in the same moment. In a productive computation setting, time is instead relevant:
\begin{lstlisting}
{ reached(Y, N + 1) } :- path(X, Y), reached(X, N), N < 10. 
\end{lstlisting}
This expression introduces a temporal aspect in going through the graph. (The boundary to \texttt{N} acts as a maximal depth of search, at the moment set as a hard-coded parameter.) We may want also to add race conditions: only one event can be caused at once (what in concurrent systems is called an \textit{interleaved semantics}). This constraint can be encoded in the following ASP rule:
\begin{lstlisting}
:- reached(X, N), reached(Y, N), X != Y.
\end{lstlisting}
Finally, the optimization criteria can be expressed as simply as:
\begin{lstlisting}
totalcost(T) :- T = #sum{C,X,Y : path(X, Y), cost(X, Y, C)}.
#minimize {T: totalcost(T)}.
\end{lstlisting}
By solving this program, we obtain the minimal path achieving all goals (colouring all target nodes) from the start conditions, together with its cost.

\section{Examples of application}

This section provides three examples of application, aiming to highlight the potential of a system like {CompLog}.

\subsection{Most relevant description}

Suppose we are given a certain ontology (including taxonomical and terminological aspects), here expressed in the form of declarative rules: 
\begin{lstlisting}
eagle -> bird. pigeon -> bird. canary -> bird.
tiger -> mammal. dog -> mammal. cat -> mammal.
dog -> pet. cat -> pet. canary -> pet.
\end{lstlisting}
Suppose we have some descriptive complexity associated to those terms (eg. related to their appearance in discourses):
\begin{lstlisting}
3 :: dog. 3 :: cat. 3 :: bird.
4 :: eagle. 4 :: tiger. 4 :: pigeon.
6 :: canary. 
\end{lstlisting}
as well as some measures on world complexity (eg. related to actual frequencies of encounter):
\begin{lstlisting}
3 :: #pigeon. 3 :: #dog. 3 :: #cat. 3 :: #bird.
7 :: #canary.
12 :: #eagle. 12 :: #tiger.
\end{lstlisting}
By computing the various measures of unexpectedness we can settle on what is the best descriptor of the current situation (minimizing $U$). For instance, given \texttt{\#pigeon}, we may prefer to say \texttt{bird} (as $C_W(\texttt{\#pigeon}) - C_D(\texttt{pigeon}) < 0$), whereas we would keep saying \texttt{eagle} for \texttt{\#eagle}.\footnote{Here we are assuming that there the rules provided with the ontology are applied with no cost. Yet, for the mental mode, they should rather be thought as reifying associationistic relations, related to how often the two concepts are activated together. If the target concept is abstract/rarely used (eg. \texttt{mammal}), additional cost may be associated to  declarative rules involving it.}

\subsection{Disjunction}

Suppose we have a die with 4 faces. In {ProbLog}, an extraction would be specified by means of an exclusive disjunction:
\begin{lstlisting}
0.25 :: die1; 0.25 :: die2; 0.25 :: die3; 0.25 :: die4.     
\end{lstlisting}
The equivalent representation in {CompLog} would distinguish the declarative from the active part (obtained eg. by program augmentation) as:
\begin{lstlisting}
2 :: die1. 2 :: die2. 2 :: die3. 2 :: die4.     
2 :: => +die1. 2 :: => +die2. 2 :: => +die3. 2 :: => +die4.
\end{lstlisting}
Note that disjunction is a by-design consequence of the race conditions. Stated otherwise, exclusive disjunction in probabilistic specifications can be seen as conveying implicitly the presence of race conditions.

\subsection{Negation}
Probability theory allows us to compute the probability of a \textit{negated event} (eg. $\neg$ \texttt{die1})  as the probability of the union set of events dual to the event which is negated, or, algebraically, as $1 - P(\texttt{die1})$.
This possibility is enabled by the closure of the horizon of events by means of an extensional semantics, ie. defining random events as sets, and then using set operations.

In contrast, Simplicity Theory (and thus {CompLog}) does not assume such a closure, as it deems cognitively unsound to infer the unexpectedness of encountering eg. a not-dog. Yet, it hints to a way to approach negation which is procedural and incremental, at a meta-level with respect to the core inference. Suppose that someone says, ``I just met an animal, which was not a dog". Plausibly, the person wants to convey that she was in a context in which she would have expected a dog, but then that expectation wasn't met. Therefore, in our mind we need first to reconstruct a context in which dog is expected, then we remove the appearance of a dog, and the first animal that would appear in its stead (eg. a cat) would give us a proxy measure for the complexity looked for. 

As a more concrete example, let us apply the same intuition to the case of the die, which is more extreme, being a fair lottery. We first compute the complexity of \texttt{die1}, which is 2. We then remove \texttt{die1} from the graph. We compute the node with the best complexity, eg. \texttt{die2}, which is still 2. We could take this as a proxy of the negation of the target, however, this computation also shows that the best alternative has the same level of unexpectedness of the target, which can be taken as a hint to proceed further. We can then apply another iteration: negating \texttt{die2}, finding \texttt{die3} with again complexity 2, and then we may aggregate the complexities of the two alternatives. This procedure can be formulated as a general heuristic: the search/aggregation can be stopped (i) when the best alternative has a sufficiently high complexity, or (ii) when the aggregated value has a sufficiently low complexity, with respect to the complexity of the target. 

\section{Conclusion and future works}
The paper presented the general motivation, the underlying theory, the main components, and elements of an initial implementation of {CompLog}, a novel complexity-based programming framework. CompLog is meant to explore a space of research in the domain of computational inferential systems (not extensional, not primarily statistical, but algorithmic, and cognitively sound inference) which has not yet attracted much attention. In future works, besides further clarifying the various components, and consolidating the various definitions, we will explore additional applications, including learning mechanisms, and attempt to integrate CompLog with existing computational frameworks, symbolic and sub-symbolic. We also aim to perform experiments to benchmark performances of CompLog on tasks where probabilistic-oriented frameworks are currently used, both in terms of computability, and of empirical alignment with human behaviour.

\bibliographystyle{splncs04}
\bibliography{references,library}

\end{document}